\documentclass[10pt,twocolumn,letterpaper]{article}
\pdfoutput=1
\usepackage{iccv}
\usepackage{times}
\usepackage{epsfig}
\usepackage{graphicx}
\usepackage{amsmath}
\usepackage{amssymb}
\usepackage{multirow}

\usepackage[breaklinks=true,bookmarks=false]{hyperref}

\iccvfinalcopy 

\def\iccvPaperID{2587} 

\ificcvfinal\fi

\begin{document}

\title{Human Trajectory Prediction via Counterfactual Analysis}

\author{%
Guangyi Chen$^{1,2,3}$, Junlong Li$^{1,2,3}$, Jiwen Lu$^{1,2,3,}\thanks{Corresponding author}$\ , Jie Zhou$^{1,2,3}$\\
{$^1$Department of Automation, Tsinghua University, China}\\
{$^2$State Key Lab of Intelligent Technologies and Systems, China}\\
{$^3$Beijing National Research Center for Information Science and Technology, China}\\
{\tt\small chen-gy16@mails.tsinghua.edu.cn; ljlong.leo@gmail.com; \{lujiwen,jzhou\}@tsinghua.edu.cn}\\
}

%

\maketitle
\ificcvfinal\thispagestyle{empty}\fi

\begin{abstract}
   Forecasting human trajectories in complex dynamic environments plays a critical role in autonomous vehicles and intelligent robots. Most existing methods learn to predict future trajectories by behavior clues from history trajectories and interaction clues from environments. However, the inherent bias between training and deployment environments is ignored. Hence, we propose a \textbf{counterfactual analysis} method for human trajectory prediction to investigate the causality between the predicted trajectories and input clues and alleviate the negative effects brought by environment bias. We first build a causal graph for trajectory forecasting with history trajectory, future trajectory, and the environment interactions. Then, we cut off the inference from environment to trajectory by constructing the counterfactual intervention on the trajectory itself. Finally, we compare the factual and counterfactual trajectory clues to alleviate the effects of environment bias and highlight the trajectory clues.
Our counterfactual analysis is a plug-and-play module that can be applied to any baseline prediction methods including RNN- and CNN-based ones. We show that our method achieves consistent improvement for different baselines and obtains the state-of-the-art results on public pedestrian trajectory forecasting benchmarks.$\footnote{Code and a video demo is available at \url{https://github.com/CHENGY12/CausalHTP}}$ 
\end{abstract}

\section{Introduction}
Human trajectory prediction aims at forecasting the future trajectory of pedestrians based on their past positions in complex and crowd environments. It is a critical and fundamental task for many applications, including the planning and controlling of the autonomous vehicles, the robot navigation, and the tracking and re-identification in the crowd surveillance. Thanks to these significances, the human trajectory prediction task has attracted much attention over the past few years~\cite{alahi2016social,xue2018ss,zhang2019sr,gupta2018social,liang2020simaug}.

\begin{figure}[t]
  \centering
  \includegraphics[width=1\linewidth]{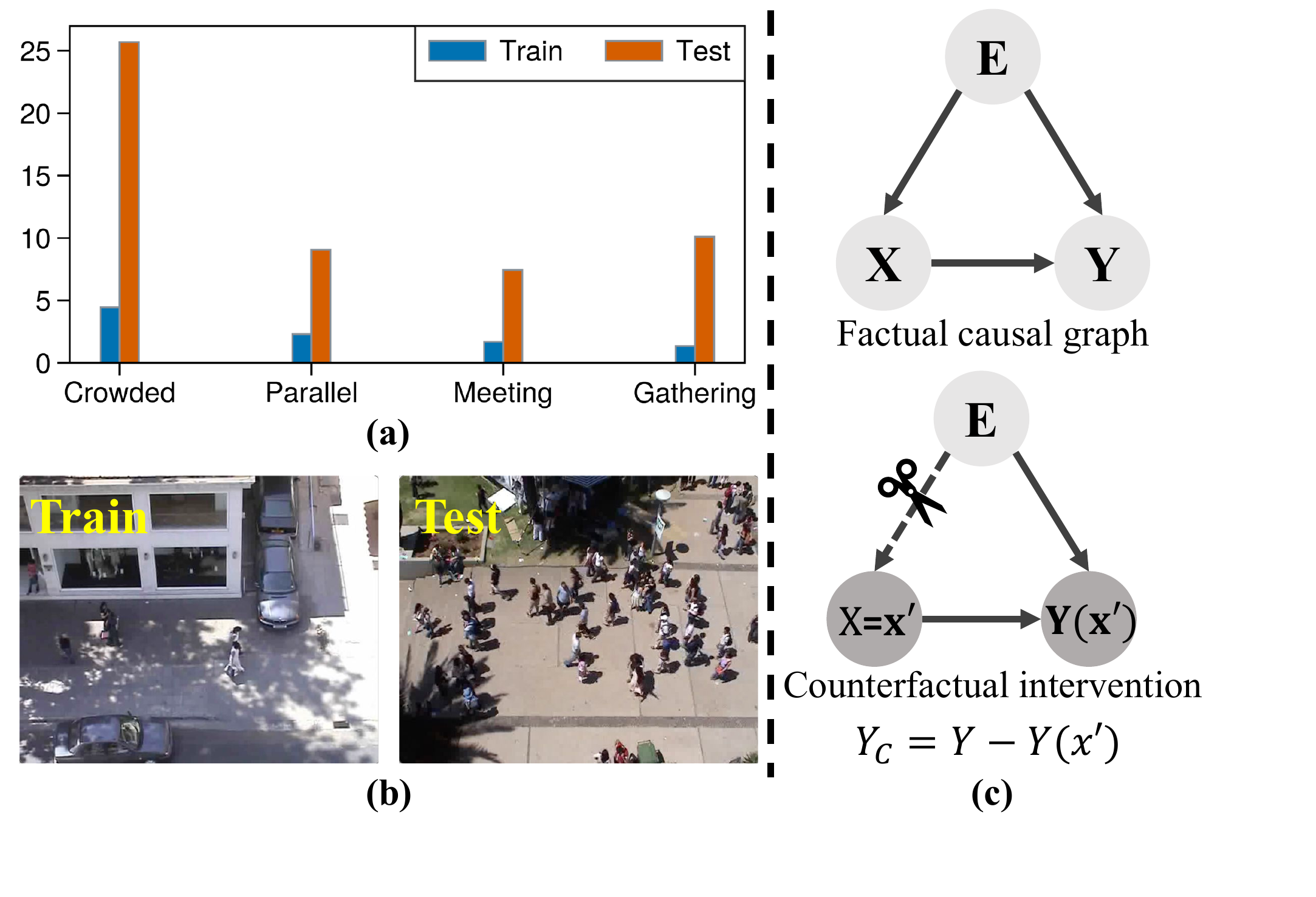}
  \caption{The motivation and key idea of our approach. In part (a), we show the bias between training and testing environments of UNIV scene. We respectively compare the degree of crowding (the average number of neighbors), the average number of parallel trajectories, the average number of trajectories that pedestrians meet, and the average number of trajectories that pedestrians gather. In part (b), we visualize the training and testing environments. We can observe obvious difference of two environments. In part (c), we first build a causal graph with the nodes E (environment interaction), X (history trajectory), and Y (future trajectory). Then, at the bottom of (c), we conduct the counterfactual intervention on the X history trajectory to cut off the dependence between history trajectory and environment interaction. The causal prediction $Y_C$ is obtained by computing the difference between the original and counterfactual predictions to alleviate the negative effects of environment bias.
}
\vspace{-0.1cm}
\label{fig:motivation}
\end{figure}

Despite the recent progress, trajectory prediction is still a challenging problem due to complex social or physical environment interactions. In the crowd environment, the pedestrian trajectory are always interacted with the social behavior from other pedestrians and the common scene context. For example, pedestrians may walk in parallel for social talking or stop for seconds to avoid collisions. Besides, the behaviors of pedestrians may be interacted by the traffic light, crosswalk, or just an obstacle like a tree.

Most existing methods concentrate on modeling the environment interactions and aggregate these interaction clues with history behavior clues for trajectory prediction.  For example, Social LSTM~\cite{alahi2016social} applies a social pooling module to extract clues from environment interaction. While Social-STGCNN~\cite{mohamed2020social} utilizes the spatial-temporal graph CNN to encode the environment interactions. 
However, these methods always ignore the inherent bias between training and deployment environments. As shown in part (a) of Figure~\ref{fig:motivation}, we statistically compare the interactions on training and testing environments of UNIV scene. We observe an obvious gap between the interactions of different environments. We also visualize the environment difference in part (b) of Figure~\ref{fig:motivation}, where the training environment is a street scene while the testing environment is a more crowded public square. This environment bias causes the heavy overfitting of complex environment interaction modules in most of trajectory prediction methods. When the training data contains many examples who turns left in the crossroads, the prediction of left trajectory will be wrongly attributed to crossroads, which misleads the prediction in the new environment where prediction turn right in the crossroads. Besides, deployment environments are always unpredictable in the applications of autonomous vehicles and intelligent robots. Thus, it is difficult to apply transfer learning methods to reduce environment bias.


To address this problem, in this paper, we propose a counterfactual analysis method to alleviate the overdependence of environment bias and highlight the trajectory clues itself. Inspired by the causal inference methods~\cite{pearl2018book,pearl2016causal,forney2017counterfactual}, we propose to use the counterfactual intervention to investigate the causality between the observed clues and predicted trajectories. Different from conventional causal inference methods~\cite{pearl2013direct,Fang_2020_CVPR}, we furthermore extend these to the training process for the optimization of the prediction model. Specifically, we first construct a factual causal graph by the human prior knowledge, whose nodes include past trajectory, environment interaction, and future trajectory. As shown in the left part of Figure~\ref{fig:motivation} (b), environment interaction may be a negative confounder due to the bias between training and deployment environments. Then, we conduct the counterfactual intervention on the history trajectory, which cuts off the dependence between the environment and trajectory. Motivated by ~\cite{vanderweele2013three,Tang_2020_CVPR}, we replace the history trajectory feature into counterfactual trajectory, such as uniform rectilinear motion, mean trajectory, or random trajectory. This counterfactual prediction indicates the effect of biased environment clues. 
Finally, we subtract the counterfactual prediction from original prediction as the causality-aware prediction since the negative effect of confounder is alleviated. We highlight that the proposed counterfactual analysis method is a plug-and-play module which can be applied to any baseline prediction method including RNN- and CNN-based ones. In the experiments, we apply our method for two baseline methods including RNN-based STGAT~\cite{huang2019stgat} and CNN-based Social-STGCNN~\cite{mohamed2020social}. We show that our counterfactual analysis method achieves consistent improvement for both two baseline methods and obtains the state-of-the-art results on public pedestrian trajectory prediction datasets.

We summarize three advantages of the proposed counterfactual analysis method as follows:
\begin{itemize}
  \item We show that the environment bias may cause the heavy overfitting of the complex environment interaction modules.
  \item We propose a counterfactual analysis method to alleviate the overdependence of environment bias and highlight the trajectory clues itself.
  \item Our counterfactual analysis is a simply plug-and-play module which can be easily applied to any baseline predictor, and consistently improves the performance on many human trajectory prediction benchmarks.
\end{itemize} 

\begin{figure*}[t]
\begin{center}
\includegraphics[width=1.0\linewidth]{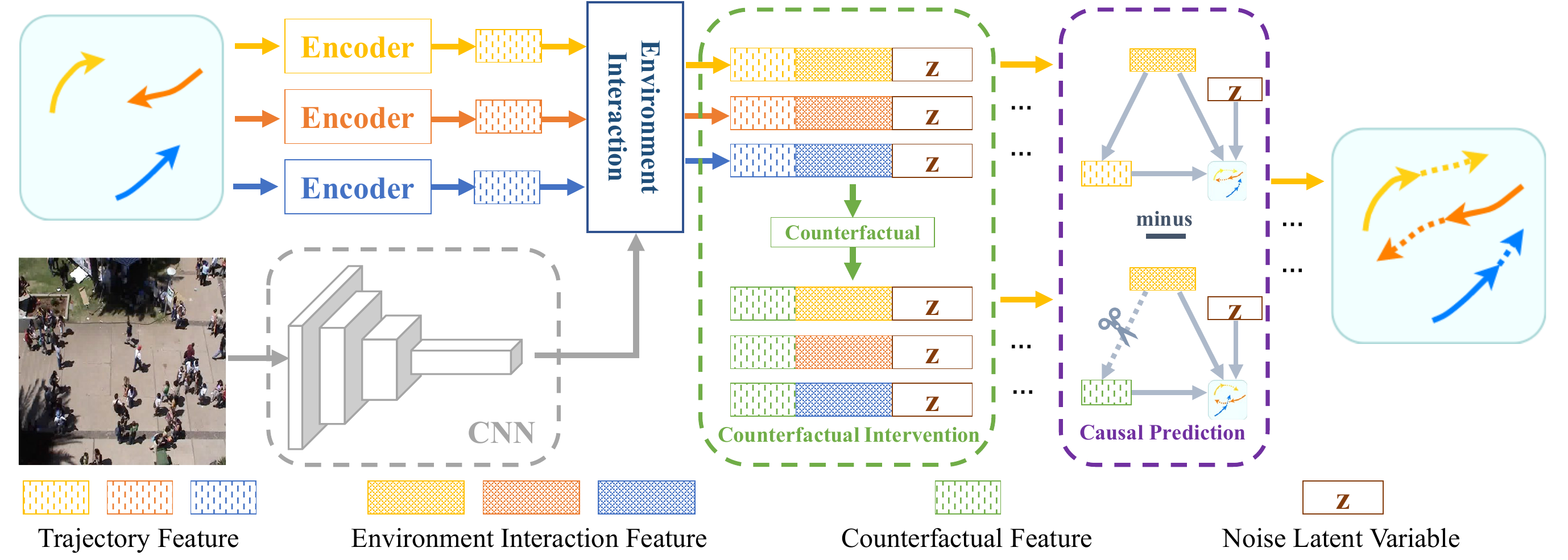}
\end{center}
   \caption{Training process of our counterfactual analysis method. We apply the counterfactual intervention by replacing the features of past trajectory with the counterfactual features such as uniform rectilinear motion (zero), mean trajectory, or random trajectory. The counterfactual prediction denotes the biased affect from environment confounder. To alleviate the negative effect of environment bias, we subtract the counterfactual prediction from original prediction as the final causal prediction.     (Best viewed in color.)}
\label{fig:network}
\vspace{-0.2cm}
\end{figure*}

\section{Related Work}
\textbf{Environment Interaction:}
The environment interaction consists of social environment interaction and physical environment interaction.
To model the complex social environment interaction, previous methods always model human crowd motion with handcrafted rules or energy parameters, e.g. Social Force Model~\cite{helbing1995social}, continuum dynamics~\cite{treuille2006continuum}, Discrete Choice framework~\cite{antonini2006discrete}, Gaussian processes~\cite{tay2008modelling,wang2007gaussian} and crowd analysis~\cite{rodriguez2011data,yamaguchi2011you,lee2016predicting}. In deep learning methods, early studies~\cite{alahi2016social,gupta2018social} employ the pooling module to capture the social interactions. While some methods~\cite{vemula2018social,fernando2018soft+,sadeghian2019sophie,yu2020spatio,chandra2019traphic} apply the attention model to distinguish the importance of different neighbors. Many recent methods employ the graph model to extract the clues from interactions, e.g. the States Refinement module for message passing in SR-LSTM~\cite{zhang2019sr}; the Graph Attention model for modeling interactions in~\cite{huang2019stgat,kosaraju2019social}; STR-GGRNN~\cite{haddad2020self} uses data-driven adaptive online neighborhood recommendation. 
In addition, some methods~\cite{lee2017desire,sadeghian2019sophie,xue2018ss,kosaraju2019social,bi2020how} also incorporate physical environment interactions of the scene context for more reliable prediction. However, most of these methods ignore the environment bias between training and deployment environments, which causes the heavy overfitting of the environment interaction modules. To address this problem, we propose a counterfactual analysis method to encourage the model to focus more on trajectory itself, instead of the biased environment interaction. 

\textbf{Predictor:}
Deep learning based methods always regard the trajectory forecasting as the sequence prediction problem and apply the RNN model~\cite{alahi2016social,gupta2018social,kosaraju2019social,zhang2019sr,sadeghian2019sophie} or temporal CNN model~\cite{nikhil2018convolutional,mohamed2020social} as the predictor. For example, Social LSTM~\cite{alahi2016social} employs the LSTM to encode human motion, then fuses the environment interaction clues extracted by a pooling module and feeds them into a LSTM decoder to sequentially predict the trajectory. While Nikhil \emph{et al.}~\cite{nikhil2018convolutional} first proposed to use the temporal CNN model to replace the RNN-based predictors. However, both RNN-based and CNN-based network are likelihood-based predictor whose causalities between observed clues and predicted results are still nontransparent. In our counterfactual analysis, we can investigate the effects of different clues including history trajectory and environment interactions by the intervention.

\textbf{Generative Modeling:} Another popular research direction of human trajectory prediction is exploring the indeterminacy of the human behavior. Many methods~\cite{gupta2018social,ivanovic2019trajectron,kosaraju2019social,zhao2019multi,huang2019stgat,tang2019multiple} employ generative model to predict multiple plausible paths, instead of the deterministic one. These methods employ Generative Adversarial Network (GAN)~\cite{goodfellow2014generative} to implicitly incorporates a latent variable into the encoded embedding ~\cite{gupta2018social,kosaraju2019social,sadeghian2019sophie,zhao2019multi}, or Variational Auto-Encoder (VAE)~\cite{kingma2014auto} to explicitly model the multi-modal distribution with a latent variable~\cite{lee2017desire,ivanovic2019trajectron,salzmann2020trajectron++,mangalam2020not}. While STGCNN~\cite{mohamed2020social} directly predicts a bi-variate Gaussian distribution to replace some deterministic trajectories with multi-modal predictions. Besides, ~\cite{giuliari2021transformer} proposes to use Transformers for predicting the future trajectories.

\textbf{Causal Inference:} Different from conventional associated inference, the causal inference aims to investigate the effect of variables when some cause is changed~\cite{pearl2016causal,pearl2009causal}. Recently, many methods take effort to apply causal inference for the effect analysis of deep neural network in different fields including reinforcement learning~\cite{nair2019causal,forney2017counterfactual}, natural language processing~\cite{vig2020causal,park2019paraphrase}, and visual representation learning~\cite{chalupka2014visual,lopez2017discovering,wang2020visual,Tang_2020_CVPR}. The clear causalities can improve the transparency of the deep models as most of them are treated as black-box nowadays.
In this paper, we propose to apply the counterfactual analysis for the trajectory prediction system to analyze the effects of different clues and enhance the reliability, which attempts to explain how neural networks make the prediction and optimize the prediction by removing the affects from biased dataset. It is helpful to understand the causal relation between the input clues and output trajectory predictions. 

\section{Approach}

\subsection{Problem Definition}

The pedestrian trajectory prediction task can be defined as a sequential prediction problem, whose inputs include the history trajectory and environment interaction. Many method designs complex model to learn the clues from environment interactions. However, the inherent bias between training and deployment environments may cause the overfitting of interaction model. In the conventional trajectory prediction framework, given N pedestrians in a scene, we can define the history trajectory of $i$th pedestrian as $X_i =\{(x_i^t,y_i^t)\in R^2| t=1,2,...,t_{obs} \} $, where the $(x_i^t,y_i^t)$ is the 2D location at time $t$. The ground-truth future trajectory of $i$th pedestrian can be defined as $Y_i =\{(x_i^t,y_i^t)\in R^2| t=t_{obs}+1,t_{obs}+2,...,t_{pred} \} $. For the normalization of scale, many methods employ the relative locations even the relative speeds to replace the absolute locations. The environment interactions can be extracted from the trajectories of other pedestrians and the physical scene context as $E_i = \mathcal{E_{\phi}}(\{ X_i \}_{i=1}^N,C) $. 
The sequential trajectory prediction process is modeled as
 \begin{equation}
  \begin{aligned}
\label{eq: prediction_orig} \hat{Y}_i = \mathcal{F}_{\theta}(X_i,E_i),
  \end{aligned}
\end{equation}
where $\hat{Y}_i =\{(\hat{x}_i^t,\hat{y}_i^t)\in R^2| t=t_{obs}+1,t_{obs}+2,...,t_{pred} \}$ denotes the predicted trajectory.
 
Given a set of training pedestrian trajectory data $\{(X_i,Y_i)\} \in \Omega $, the predictor is optimized by the L2 loss function as:
\begin{equation}
  \begin{aligned}
\label{eq: likelihood_training} \mathcal{L}_{tri}(\theta,\phi|\Omega)= L_{L2}(Y_i,\hat{Y_i}),
  \end{aligned}
\end{equation}
where $\theta,\phi$ denote the parameters of the pedestrian trajectory prediction system.

\subsection{Causal Graph}

In this subsection, we reformulate above trajectory prediction framework with the causal graph $\mathcal{G}=\{\mathcal{V},\mathcal{E} \}$ with the prior knowledge. As shown in part (c) of Figure~\ref{fig:motivation}, the nodes $\mathcal{V}$ in the graph denote the variables including history trajectory $X$, environment interactions $E$ (including social and physical environment interactions), and future trajectory $Y$. While the causal links $\mathcal{E}$ indicate the hidden causal relations and how these variables interact with each other. 
In a link $E\rightarrow X \rightarrow Y $, we can call the node $E$ as the parent node of $X$, while $Y$ is the child node of $X$. The link from the parent node to the child node denotes a causality. $E \rightarrow (X,Y) $ indicates all the human trajectories are influenced by the environment. In the causal inference theory~\cite{pearl2009causality}, when a variable ($E$) simultaneously affects two variables, it becomes a confounder in the causal analysis of these two variables. For example, if turning left is often observed with the environment interaction of crossroads, the prediction of left trajectory will be wrongly attributed to crossroads. While the real causality underlying the history trajectory may be ignored.


\subsection{Counterfactual Analysis}

For conventional likelihood-based predictors, the causalities between different observed clues and predicted results are nontransparent. The prediction model is easy to be ``cheat" by the short cut between the biased environment interaction and final future trajectory. Inspired by the causal inference methods~\cite{pearl2009causality,pearl2013direct} which attempt to analyze the causalities among different clues, we propose to apply the counterfactual analysis for trajectory prediction system to mitigate the negative effect of biased environment and encourage the model to focus more on trajectory itself. Counterfactual intervention means to imagine a nonexistent observation to replace the original factual clues in the trajectory prediction system. In the causal inference methods~\cite{pearl2018book}, the intervention is formulated as $do(\cdot)$. Once one variable is intervened, its all in-coming links in the causal graph are cut off and its value is independently given, while other variables who are not affected still maintain the original value. In our method, as shown in part (c) of Figure~\ref{fig:motivation}, we replace the history trajectory features in the system with the imagined trajectory or its embedding. We can obtain the prediction results under this intervention as:
 \begin{equation}
  \begin{aligned}
\label{eq: prediction_counterfactual} \hat{Y}_{X_i=x'}= \mathcal{F}_{\theta}(do(X_i=x'),S_i,I),
  \end{aligned}
\end{equation}
where $X=x'$ indicates the counterfactual value. Specifically, we can apply the different counterfactual interventions such as the zero vector (which denotes uniform rectilinear motion when the relative speed is used as input), the mean vector of all history trajectories or the random trajectories as the counterfactual intervention.

The original factual prediction $\hat{Y}_i$ is dependent on both trajectories and environment interactions, while the counterfactual predicted result $\hat{Y}_{X_i=x'} $ is only dependent with environment interactions since the trajectory is replaced by counterfactual intervention.
To investigate the real effect of trajectory itself, we compute the difference between the counterfactual prediction and factual prediction and name it as the causal prediction:
 \begin{equation}
  \begin{aligned}
\label{eq: prediction_counterfactual2} \hat{Y}_{causal} = \hat{Y}_i- \hat{Y}_{X_i=x'}.
  \end{aligned}
\end{equation}
Compared with original likelihood prediction, the causal prediction is more reliable by avoiding the biased affects from environment confounders. 
In the training process, we optimize the network to predict the causal prediction:
 \begin{equation}
  \begin{aligned}
\label{eq: causal_training} \mathcal{L}_{Causal}(\theta,\phi|\Omega)= L_{L2}(Y_i,\hat{Y}_{causal}).
  \end{aligned}
\end{equation}

\subsection{Generative Model}
It is worth noting that our counterfactual analysis is robust to the slight change of trajectory prediction systems, such as different predictors, whether the physical environment content is used, or whether generative model is used. These changes cannot affect the structure of the causal graph.
When using the generative methods, we will add the noise latent variable in the trajectory prediction system and predict the stochastic future trajectory. Taking the GAN as the example of generative model, the loss function in \eqref{eq: likelihood_training} is reformulated as 
\begin{equation}
  \begin{aligned}
\label{eq: gan_training} \mathcal{L}_{GAN}(\theta,\phi,\psi|\Omega)&= L_{L2}(Y_i,\hat{Y_i}) + log D_{\psi}(Y_i) \\ & + log(1-D_{\psi}(\hat{Y_i})),
  \end{aligned}
\end{equation}
where  $\hat{Y_i} =\mathcal{F}_{\theta}(X_i,E_i,Z) $ indicates the trajectory is generated by a noise latent variable $Z$ for multi-modal distributions.
 While $D_{\psi}$ is the discriminator in the GAN to judge whether the trajectory is generated by the noise latent variable. Our counterfactual analysis can work together with generative methods, which only needs to add the noise latent variable $Z$ in the side of predicted trajectory $Y$ with $Z\rightarrow Y $.
The causal prediction with the latent variable is formulated as: 
\begin{equation}
  \begin{aligned}
\label{eq: prediction_counterfactual_gan} \hat{Y}_{causal} &= \hat{Y}_i(X_i,E_i,Z )\\&- \hat{Y}_{X_i=x'}(do(X_i=x'),E_i,Z ).
  \end{aligned}
\end{equation}
Then, we only need to replace the original prediction with causal prediction as:
\begin{equation}
  \begin{aligned}
\label{eq: gan+causal_training} \mathcal{L}_{CausalGAN}(\theta,\phi,\psi|\Omega)&= L_{L2}(Y_i,\hat{Y}_{causal}) + log D_{\psi}(Y_i) \\&+ log(1-D_{\psi}(\hat{Y}_{causal})).
  \end{aligned}
\end{equation}


\subsection{Implementation Details}

As shown in Figure~\ref{fig:network}, the framework of trajectory prediction system always contains 4 modules including history path encoder, scene perceptron, interactions analysis module, and trajectory predictor. Instead of the conventional likelihood-based trajectory predictor, in this paper, we conduct the counterfactual intervention for some clue and apply a causal predictor by computing the difference between the prediction results with factual and counterfactual clues. Our causal prediction is simple yet effective for the reliability of trajectory prediction system. To evaluate the generality of our counterfactual analysis method, we incorporate it as an plug-and-play module into two different baseline methods including RNN-based STGAT~\cite{huang2019stgat} and CNN-based Social-STGCNN~\cite{mohamed2020social}. We simply introduce these two implementations as follows and the details can be found in the supplementary materials. Note that though the whole counterfactual analysis method contains the features of environment scene, here we \textbf{do not} use the scene features in the implementations since our baseline models do not use scene images. Additionally, we conduct experiments \textbf{with} the scene features in the appendix.

\begin{table*}[t]
\caption{Comparison with several state-of-the-art models. We notice that our causal-based methods have the best average error for RNN-based and CNN-based methods. The lower is the better.}
\begin{center}
\begin{tabular}{l|c|c|c|c|c|c}
\hline
\multirow{2}*{RNN-based Method}  &  \multicolumn{6}{c}{Performance (ADE/FDE)} \\
\cline{2-7}
~ & ETH & HOTEL &  ZARA1&  ZARA2 &UNIV & AVG \\
\hline
LSTM  & 1.09/2.41     & 0.86/1.91     & 0.41/0.88     & 0.52/1.11     & 0.61/1.31     & 0.70/1.52 \\
S-LSTM~\cite{alahi2016social}  & 1.09/2.35     &0.79/1.76    & 0.47/1.00     & 0.56/1.17     & 0.67/1.40     &  0.72/1.54 \\
SGAN~\cite{gupta2018social}   & 0.81/1.52     & 0.72/1.61     & 0.34/0.69     & 0.42/0.84     & 0.60/1.26     &  0.58/1.18 \\
Sophie~\cite{sadeghian2019sophie}  & 0.70/1.43     & 0.76/1.67     & 0.30/0.63   & 0.38/0.78     & 0.54/1.24     & 0.54/1.15 \\
SR-LSTM~\cite{zhang2019sr}  & {0.63/1.25}   & 0.37/0.74     & 0.41/0.90     & {0.32/0.70}   & 0.51/1.10    & {0.45/0.94} \\
Social-BiGAT~\cite{kosaraju2019social}  & 0.69/1.29     & 0.49/1.01     & 0.30/0.62     & 0.36/0.75     & 0.55/1.32     & 0.48/1.00 \\
MATF~\cite{zhao2019multi}   & 1.33/2.49      & 0.51/0.95     & 0.44/0.93     & 0.34/0.73     & 0.56/1.19     & 0.64/1.26 \\
MATF GAN~\cite{zhao2019multi}  & 1.01/1.75   & 0.43/0.80     &  0.26/\textbf{0.45}    & 0.26/0.57     & 0.44/0.91     & 0.48/0.90  \\
IDL~\cite{yamaguchi2011you}     & \textbf{0.59}/1.30     & 0.46/0.83     & \textbf{0.22}/0.49     & \textbf{0.23}/\textbf{0.55}     & 0.51/1.27             & \textbf{0.40}/0.89 \\
PIF~\cite{liang2019peeking} & 0.73/1.65  & \textbf{0.30}/0.59 & 0.60/1.27 & 0.38/0.81 & \textbf{0.31}/\textbf{0.68} & 0.46/1.00 \\
STGAT~\cite{huang2019stgat}    & 0.65/1.12     & 0.35/0.66     & 0.34/0.69     & 0.29/0.60     & 0.52/1.10     & 0.43/0.83 \\
\hline
STGAT$^*$ &0.73/1.39     & 0.38/0.72     & 0.35/0.69     & 0.32/0.64     & 0.57/1.22     & 0.47/0.93 \\
\textbf{Causal-STGAT} &  0.60/\textbf{0.98}  & \textbf{0.30}/\textbf{0.54}     & 0.32/0.64     & 0.28/0.58     & 0.52/1.10    & \textbf{0.40}/\textbf{0.77} \\

\hline\hline
\multirow{2}*{CNN-based Method}  &  \multicolumn{6}{c}{Performance (ADE/FDE)} \\
\cline{2-7}
~ & ETH & HOTEL &  ZARA1&  ZARA2 &UNIV & AVG \\
\hline
CNN~\cite{nikhil2018convolutional} & 1.04/2.07 & 0.59/1.17     & 0.43/0.90     & 0.34/0.75     & 0.57/2.32     & 0.59/1.22 \\
Social-STGCNN~\cite{mohamed2020social} & \textbf{0.64}/1.11     & 0.49/0.85     & \textbf{0.34}/\textbf{0.53}     & \textbf{0.30}/0.48     & \textbf{0.44}/\textbf{0.79}     & 0.44/0.75 \\
\hline
Social-STGCNN$^*$ & 0.66/1.12 & 0.40/0.65     & 0.35/0.56     & 0.31/\textbf{0.47}     & 0.51/0.93     & 0.45/0.75 \\
\textbf{Causal-STGCNN} &  \textbf{0.64}/\textbf{1.00}  & \textbf{0.38}/\textbf{0.45}     & \textbf{0.34}/\textbf{0.53}     & 0.32/0.49     & 0.49/0.81    & \textbf{0.43}/\textbf{0.66} \\
\hline
\end{tabular}
\end{center}
\label{table:sota}
\vspace{-0.1cm}
\end{table*}

\begin{table*}[t]
\caption{Evaluation of Causal-STGAT for conducting different counterfactual interventions on different clues.}
\vspace{0.1cm}
\begin{center}
\begin{tabular}{l|c|c|c|c|c|c}
\hline
\multirow{2}*{Method}  &  \multicolumn{6}{c}{Performance (ADE/FDE)} \\
\cline{2-7}
~ & ETH & HOTEL &  ZARA1&  ZARA2 &UNIV & AVG \\
\hline
STGAT$^*$ (baseline) &0.73/1.39     & 0.38/0.72     & 0.35/0.69     & 0.32/0.64     & 0.57/1.22     & 0.47/0.93 \\
\hline
Causal-STGAT-Zero &  0.60/0.98  & 0.30/0.54     & 0.32/0.64     & 0.28/0.58     & 0.52/1.10    & 0.40/0.77 \\
Causal-STGAT-Mean &  0.65/1.17 & 0.33/0.60  & 0.35/0.71 & 0.29/0.60  & 0.54/1.15 & 0.44/0.84   \\
Causal-STGAT-Random &  0.59/1.05 & 0.31/0.55 &  0.34/0.70 & 0.28/0.59 & 0.53/1.13 & 0.42/0.80   \\
\hline
\end{tabular}
\end{center}
\label{table: implementation}
\vspace{-0.1cm}
\end{table*}

\begin{table*}[!t]
\centering
\caption{Parameters size and inference time of baseline models compared to ours. The lower is better.}
\vspace{0.1cm}
\label{table: speed}
\begin{center}
\begin{tabular}{c|c|c|c|c}
    \hline
    Method& Social-STGCNN & Causal-STGCNN & STGAT & Causal-STGAT\cr
    \hline
    Inference Speed & 0.0116 & 0.0124  & 0.3343 & 0.3418\cr
    \hline
    Parameters Count & 7.6k & 7.6k & 56k & 56k \cr
      \hline
\end{tabular}
\end{center}
\vspace{-0.2cm}
\end{table*}

\textbf{Causal-STGAT:} The STGAT method employs two LSTM and a graph attention network (GAT) to encode the history trajectory and social interaction clues. Specifically, M-LSTM (LSTM for motion coding) focuses on the history trajectory features, while G-LSTM (LSTM for graph) and GAT models are applied to extract the interaction features. Then the features from M-LSTM, G-LSTM + GAT and latent noise are connected as the input of the LSTM predictor. For example about history trajectory, we replace the feature vector from M-LSTM with the counterfactual intervention (the mean vector of all trajectory features, or the zero vector). Then we respectively conduct predictions with original connected feature and counterfactual connected feature whose history trajectory part is changed. Finally, we adopt the difference between two results as the prediction of our Causal-STGAT.

\textbf{Causal-STGCNN:} Social-STGCNN employs the spatial-temporal graph convolution neural network (ST-GCNN) as the encoder to extract the features from history trajectory and social interactions. The trajectories in a scene are represented as a graph $G=\{N,E\} $, where nodes $N$ denote the past trajectories, while the adjacency matrix $A$ in the GCN denotes the social interactions. Then a time-extrapolator CNN is applied as the trajectory predictor with the outputs from ST-GCNN. In this baseline, we replace the original nodes $N$ by counterfactual intervention and maintain the values of interaction clues $A$. Then we respectively feed the outputs of ST-GCNN with factual and counterfactual nodes to the time-extrapolator CNN predictor and compute the difference to get the causal prediction.

\section{Experiments}

\subsection{Datasets and Experimental Settings}

\textbf{Datasets:} We conduct experiments on two publicly available human trajectory prediction datasets: ETH~\cite{pellegrini2010improving} and UCY~\cite{lerner2007crowds}. ( Besides, we provide experimental results on Stanford drone dataset~\cite{robicquet2016learning} in the appendix. ) The human trajectories in both two datasets are captured in the real world scenes with rich social interactions.  These datasets contain five unique scenes: Zara1, Zara2, Univ, ETH, and Hotel with 1536 detected pedestrians. All trajectories in datasets are sampled every 0.4 seconds. The experimental settings follow the previous methods like Social-GAN~\cite{gupta2018social}, Social BiGAT~\cite{kosaraju2019social},and STGAT~\cite{huang2019stgat}. We also use the leave-one-out approach to train and validate our model on 4 sets and test on the remaining one. During evaluation, the first 3.2 seconds (8 frames) are observed and the next 4.8 seconds (12 frames) are to be predicted.


\begin{figure*}[t]
\begin{center}
\includegraphics[width=1.0\linewidth]{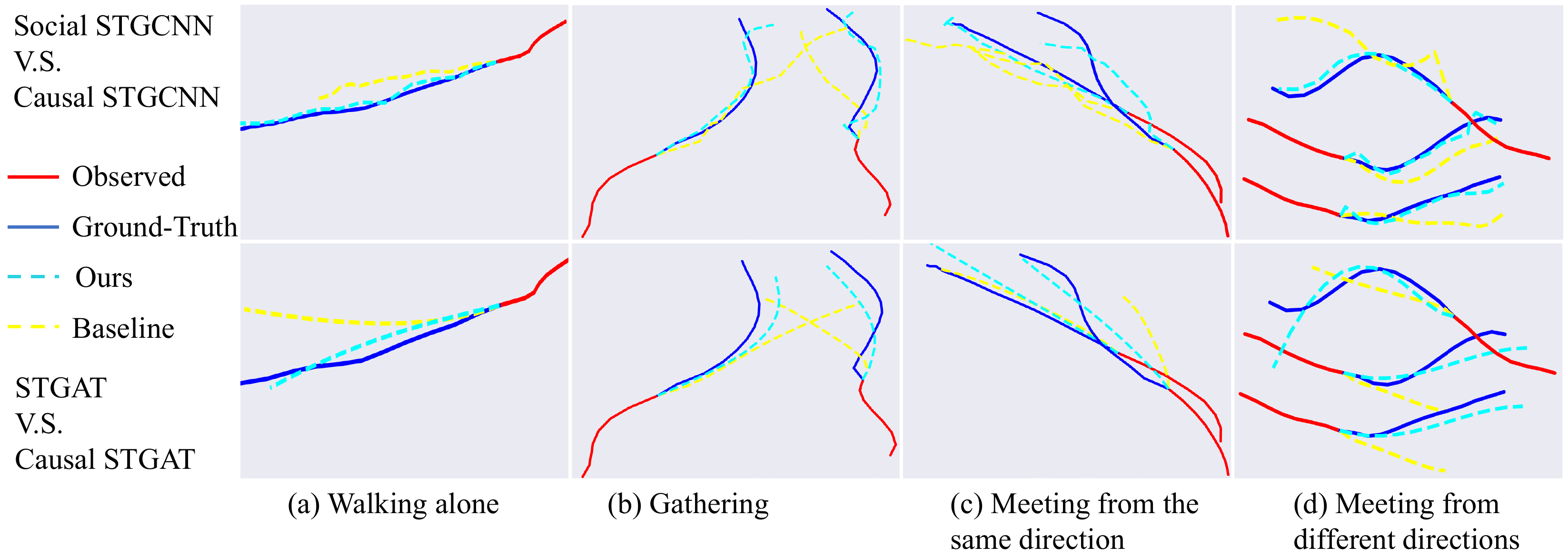}
\end{center}
   \caption{ Qualitative analysis for our causal prediction methods. We compare our causal prediction methods (Causal-STGCNN and Causal-STGAT) with corresponding baselines in four scenarios: (a) a person walking alone; (b) two pedestrians gathering; (c) two persons meeting from the same direction; (d) two persons meeting from different directions.}
\label{fig:show trajectory}
\end{figure*}

\textbf{Evaluation Metrics:} We adopt the same evaluation metrics as prior methods~\cite{gupta2018social,kosaraju2019social,huang2019stgat}, including Average Displacement Error (ADE) and Final Displacement Error (FDE). ADE computes the mean square error (MSE) of predicted trajectory and ground-truth trajectory, while FDE computes the L2 distance between the final locations of predicted and ground-truth trajectories. Since both baselines Social-STGCNN~\cite{mohamed2020social} and STGAT~\cite{huang2019stgat} are generative methods, we also follow the evaluation method in Social-GAN~\cite{gupta2018social}. We generate 20 samples based on the predicted distribution and select the closest sample to the ground truth for ADE and FDE metrics.

\begin{figure*}[t]
  \centering
  \includegraphics[width=0.99\linewidth]{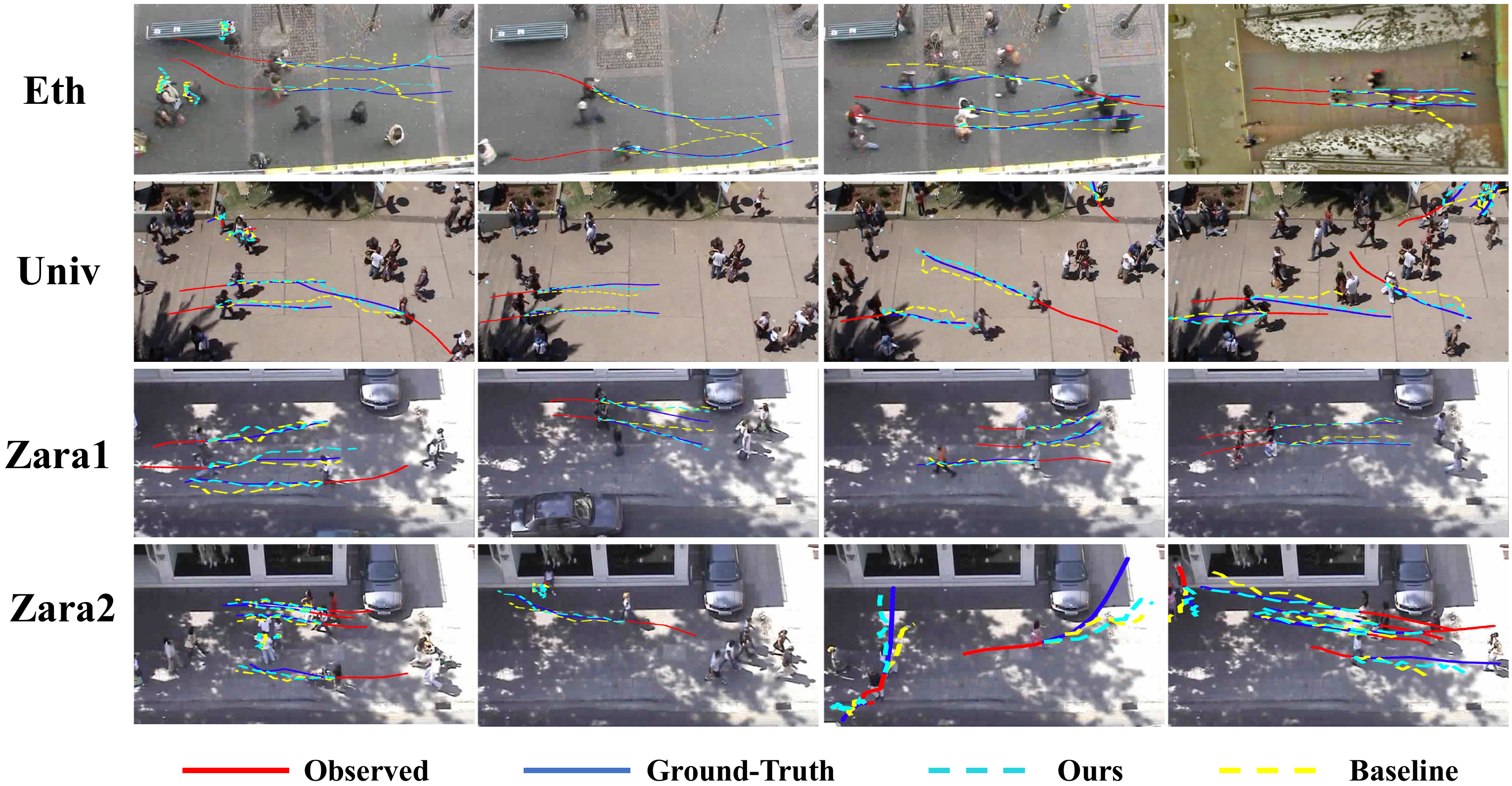}
  \caption{ Visualization examples of our Causal-STGCNN method in different scenes. We compare the trajectories generated by our Causal-STGCNN method and baseline Social-STGCNN method in the ETH and UCY datasets. Our causal-based method can solve the complex scenes and social interactions with multiple humans by the counterfactual analysis. (Best viewed in color.)
}
\label{fig:examples_cnn}
\end{figure*}

\begin{figure*}[t]
  \centering
  \includegraphics[width=0.99\linewidth]{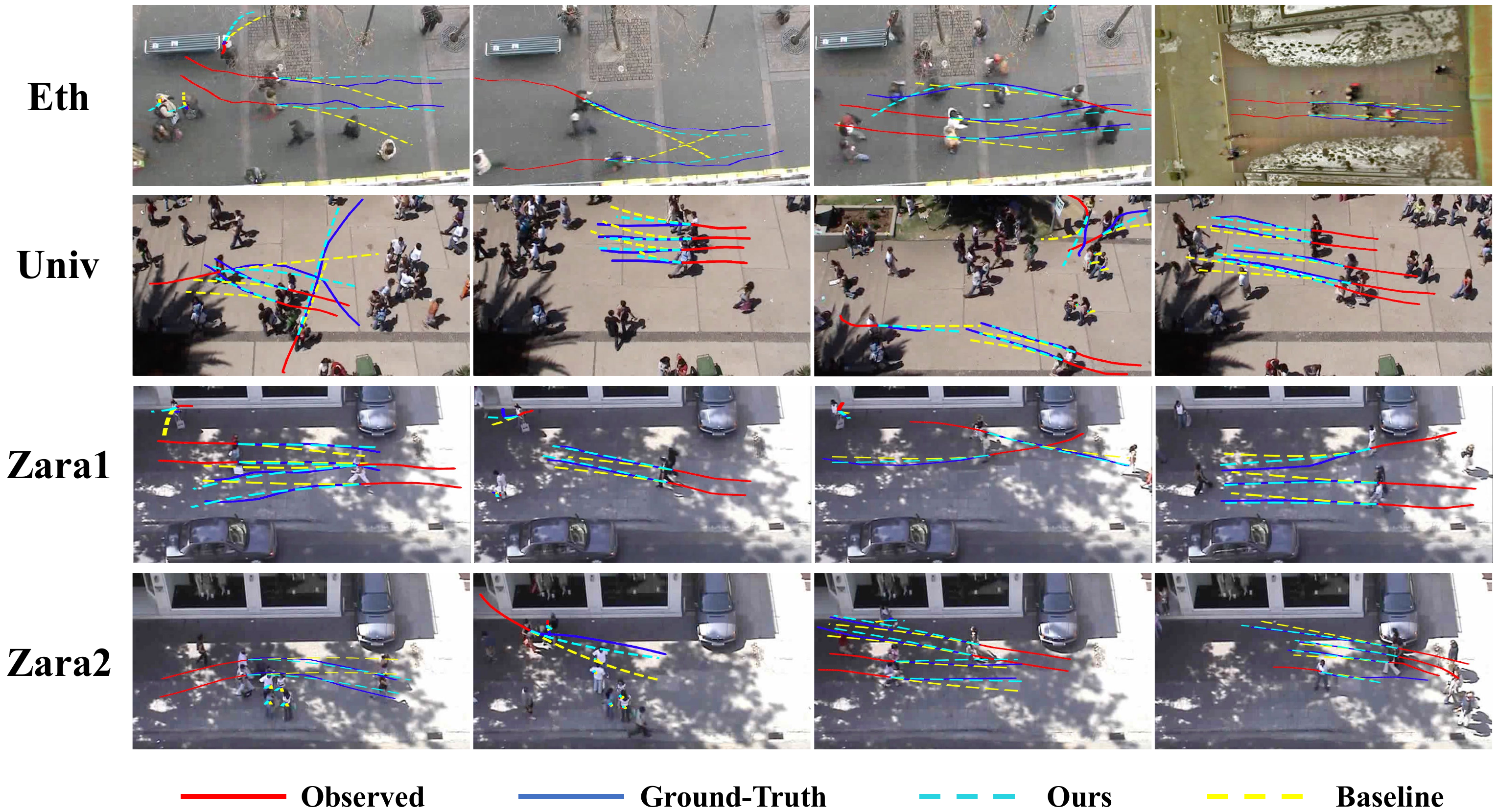}
  \caption{ Visualization examples of our Causal-STGAT method and baseline Social-STGAT method in the different scenes in the both ETH and UCY datasets. The comparisons quantitatively demonstrate the effectiveness of our counterfactual analysis on the RNN-based baselines. (Best viewed in color.)
}
\label{fig:examples_gat}
\end{figure*}

\subsection{Quantitative Evaluation}

\textbf{Evaluation of Counterfactual Analysis:} 
To verify the effectiveness of our method, we compare the performance of our Causal-STGAT, Causal-STGCNN with the RNN-based baseline STGAT~\cite{huang2019stgat} and CNN-based Social-STGCNN~\cite{mohamed2020social} methods. The performance comparison about our causal-based methods, baseline methods, and other SOTA methods is summarized as Table~\ref{table:sota}. Note that the $*$ in Table~\ref{table:sota} means our reproduced results trained with the official released code. $\footnote{We reproduced the STGAT~\cite{huang2019stgat} methods with the released code from \url{https://github.com/huang-xx/STGAT}, while the code of Social-STGCNN~\cite{mohamed2020social} comes from \url{https://github.com/abduallahmohamed/Social-STGCNN}. }$ Due to different implementation environments, the performance of our reproduced baselines are slightly lower then the performance reported in the paper. For the fair comparison, all hyper-parameters and environments of our Causal-STGAT and Causal-STGCNN methods are same with the reproduced baselines. 
As shown in Table~\ref{table:sota}, our causal prediction method can consistently improve the performance than the original likelihood-based prediction with different baselines on most of datasets. Specifically, Causal-STGAT obtains the +0.07/+0.16 ADE/FDE improvement than the baseline STGAT method on the average of 5 scenes, while Causal-STGCNN outperforms the reproduced Social-STGCNN method by over +0.02/+0.09 ADE/FDE results. 

Besides, we also compare the proposed causal based method with other SOTA methods, such as IDL~\cite{yamaguchi2011you}, Social-BiGAT~\cite{kosaraju2019social} and MATF~\cite{zhao2019multi}. As shown in Table~\ref{table:sota}, we achieves the SOTA performance with slight improvement over the RNN-based IDL~\cite{yamaguchi2011you} method and CNN-based Social-STGCNN~\cite{mohamed2020social} method. It also demonstrates the effectiveness of the proposed counterfactual analysis method. 
For both causal method in Table~\ref{table:sota}, we conducted the counterfactual intervention on the history trajectory by replacing the features with zero vector. Other attempts are introduced below.


\textbf{Evaluation of Different Counterfactual Implementation:} We have attempted different implementations of counterfactual intervention, including using the zero vector, the mean of all feature vector or a random vector sampled from a uniform distribution with $[-0.1,0.1]$. These intervention implementations (``Zero'', ``Mean'', ``Random'') with limited computing cost are usually adopted~\cite{tang2020unbiased} for causal model. For the zero vector and mean vector, the counterfactual intervention in the training and testing stage is invariant. While for random vector, we sampled the vector from a uniform distribution in the training stage and using the expectation of uniform distribution (zero vector) in the testing stage to avoid introducing the bias for testing data. As shown in bottom part of Table~\ref{table: implementation}, all implementations obtain obvious improvements for the baselines, which demonstrates the generality of the proposed counterfactual analysis method. Furthermore, 1) The ``Zero'' obtains slightly better performance than the others. It might be because the ``Zero'' is a harder counterfactual intervention, which better emphasizes the causal effect from history behaviors to future trajectories. 2) The performance of these implementations is close, which demonstrates the robustness of the implementation of counterfactual intervention. All  versions of counterfactual intervention are conducted on the human past trajectory clue.

\textbf{Evaluation of Inference Speed and Model Size:} Model size and inference speed are also critical for the deployment of the method in the real-world environment. We evaluated the speed of all models with all data in five scenes and compute the average inference time for one trajectory on one GTX 2080Ti GPU. For fairness, we fixed the batch-size as 1 for all methods and repeat the evaluation 3 times for average. The parameters cost and inference speed of our methods and baselines are summarized in Table~\ref{table: speed}. First, our counterfactual analysis method does not need any extra parameters since all parameters are shared with factual and counterfactual parts. Second, there is no such thing as a free lunch. The speed of our methods are lower than the baselines because of the extra computations for counterfactual analysis. However the extra speed cost is not heavy since the counterfactual analysis only uses part of network.

\subsection{Qualitative Evaluation} 

We qualitatively analyze how our causal predictions are more reliable than conventional likelihood-based predictions. As shown in Figure~\ref{fig:show trajectory}, we analyze our methods in 4 different scenarios. Taking the (b) as an example, given two trajectories gathering, the conventional likelihood-based methods always predict they will meet each other since the negative effect brought by the training bias of the environment interaction, (like ``most trajectories getting closer finally meet"). These training biases mislead the predictor to learn the spurious correlations of environments and ignore the real clues from history trajectory, (like ``these persons tend to gather and move forward together"). By counterfactual analysis, we remove the negative effect contained in the counterfactual prediction from original predictions. It is effective to overcome the training bias and encourage the model to highlight the real causation itself.

Besides, we also provide the visualization examples with the environment scenes. As shown in Figure~\ref{fig:examples_cnn} and Figure~\ref{fig:examples_gat}, we respectively compare our Causal-STGCNN and Causal-STGAT methods with corresponding baselines in different scenes. We observe that our counterfactual analysis method can effectively captures the real causal relations instead of the biased environment interactions. Taking the second scene in ETH as an example, both CNN-based and RNN-based counterfactual analysis methods significantly outperform the baseline methods and generate more reliable predicted trajectories.

\section{Conclusion}

In this paper, we have presented a counterfactual analysis method to investigate the causalities between the observed clues and predicted trajectories.
We apply the counterfactual intervention by replacing the features of trajectory clues with the counterfactual one, and subtract this counterfactual prediction from original prediction. By this comparison, we encourage the model to learn the real causations of trajectory and alleviate the negative effects brought by the bias between training and deployment environments.
 Our counterfactual analysis is a plug-and-play module, which can be employed for off-the-shell trajectory prediction models. In the experiments, we demonstrate the effectiveness of our counterfactual analysis method for different scenes, analyze the effects of different counterfactual implementations, and evaluate the generalization ability for different baseline methods.

{\small
\bibliographystyle{ieee_fullname}
\bibliography{trajectory}
}

\clearpage
\begin{appendix}
\section{Physical Environment}
We conduct experiments for physical environment and evaluate whether our causal model can be used for physical environment. We added a visual feature branch to the original Social-STGCNN method as the baseline. The visual feature is extracted by ResNet34~\cite{he2016deep}, and we concatenate it with the node feature and position embedding after an MLP. Then, we applied our causal model to it to mitigate the effect of biased physical and social environment. 
As shown in Table~\ref{tab: physical}, our method achieves very significant improvement by applying the causal model to the physical and social environment.

\renewcommand\tabcolsep{1pt}
\begin{table}[h]
\footnotesize
\begin{center}
\caption{Evaluation for the physical environment. (S-STGCNN denotes using scene image, C denotes our causal model)  }
\begin{tabular}{c|cccccc}
\hline
Dataset		& ETH & HOTEL &  ZARA1&  ZARA2 & UNIV & AVG \\
\hline
S-STGCNN   & 0.80/1.39 & 0.68/0.91  & 0.36/0.58 & 0.39/0.51 & 0.62/1.17 & 0.57/0.91\\ 
S-STGCNN-C & 0.65/1.18 & 0.41/0.62  & 0.34/0.56 & 0.31/0.50 & 0.45/0.80 & 0.43/0.73\\
\hline
\end{tabular}
\label{tab: physical}
\end{center}
\end{table}
\vspace{-20pt}

\section{Experiment on SDD}
The Stanford drone dataset (SDD)~\cite{robicquet2016learning} is a well established human trajectory prediction benchmark, consisting of 20 scenes and over 11, 000 unique pedestrians. It provides the scenes in bird's eye view and the locations of agents in pixel co-ordinates. More than 40, 000 interactions between the agent and scene, and over 185, 000 interactions between agents[5] are captured in the dataset\cite{robicquet2016learning}. As used in \cite{mangalam2020not, gupta2018social, sadeghian2019sophie}, we use the standard test train split for the experiments on SDD.

For the SDD dataset, we use PECNet~\cite{mangalam2020not} as our baseline and apply our causal model to it. Similar to Causal-STGAT and Causal-STGCNN, we replace the trajectory feature vector from the past trajectory encoder with counterfactual intervention (the zero vector), after which we respectively use the original feature and counterfactual feature for both destination prediction and social pooling. Then, we adopt the difference between two outputs as the causal pooled feature, and finally use it to yield the prediction of our Causal-PECNet.

\begin{table}[h]
\footnotesize
\begin{center}
\caption{Comparison with several state-of-the-art models on the SDD dataset.}
\begin{tabular}{c|ccc|cc}
\hline
Method & CF-VAE~\cite{bhattacharyya2019conditional} & SimAug~\cite{liang2020simaug} & PECNet~\cite{mangalam2020not} & PECNet$^*$ & Causal-PECNet  \\
\hline
ADE   & 12.60 & 10.27 & 9.96 & 10.27 & \textbf{9.19} \\ 
FDE & 22.30 & 19.71 & 15.88 & 15.99 & \textbf{15.31} \\
\hline
\end{tabular}
\label{tab: PEC}
\end{center}
\end{table}
\vspace{-20pt}

The $*$ in Table~\ref{tab: PEC} means our reproduced results trained with the official released code. In the table, we show the reported performance of PECNet, our reproduced performance of PECNet, and Causal-PECNet (applying our method to PECNet). Our method obtains improvement over the SOTA PECNet baseline on SDD, which demonstrates its effectiveness.

\end{appendix}

\end{document}